\PassOptionsToPackage{table}{xcolor}
\documentclass[10pt,twocolumn,letterpaper]{article}
\usepackage[pagenumbers]{cvpr}
\usepackage[T1]{fontenc}
\usepackage{makecell}
\usepackage{tabularx}
\usepackage{balance}
\usepackage{url}
\usepackage{tcolorbox}
\usepackage{multirow}
\tcbuselibrary{breakable,skins,listings}

\tcbset{reframeprompt/.style={
  enhanced,
  breakable,
  colback=gray!6,
  colframe=gray!65!black,
  colbacktitle=gray!65!black,
  coltitle=white,
  fonttitle=\small\bfseries\raggedright,
  fontupper=\small,
  boxrule=0.45pt,
  arc=2pt,
  left=8pt,right=8pt,top=6pt,bottom=6pt,
  before skip=9pt,after skip=9pt,
  pad at break*=2mm
}}

\newtcolorbox{promptbox}[2][]{reframeprompt,title={#2},#1}
\newtcblisting{promptcode}[2][]{
  reframeprompt,
  title={#2},
  listing only,
  listing options={
    basicstyle=\ttfamily\footnotesize,
    columns=fullflexible,
    keepspaces=true,
    showstringspaces=false,
    upquote=true,
    breaklines=true,
    breakatwhitespace=false,
    breakindent=0pt,
    aboveskip=0pt,
    belowskip=0pt,
    escapeinside={(*@}{@*)}
  },
  #1
}

\definecolor{imagegold}{RGB}{184,107,35}
\definecolor{videoblue}{RGB}{0,135,175}
\definecolor{dependencepurple}{RGB}{125,100,165}
\definecolor{toolteal}{RGB}{40,140,110}
\newcommand{\imagescoregain}[2]{#1\textsubscript{\textcolor{imagegold}{#2}}}
\newcommand{\videoscoregain}[2]{#1\textsubscript{\textcolor{videoblue}{#2}}}
\newcommand{\gain}[1]{\textcolor{green!50!black}{#1}}
\newcommand{\regression}[1]{\textcolor{red!75!black}{#1}}
\newcommand{\cmark}{\textcolor{green!55!black}{\ensuremath{\checkmark}}}
\newcommand{\xmark}{\textcolor{red!80!black}{\ensuremath{\times}}}

\definecolor{cvprblue}{rgb}{0.21,0.49,0.74}
\usepackage[breaklinks,colorlinks,allcolors=cvprblue]{hyperref}
\hypersetup{pdftitle={Reframe: Observation-Aligned Multi-Teacher On-Policy Distillation for Agentic Visual Reasoning}, pdfauthor={Haozhi Yuan, Handong Li, Yifan Zhang, Qianshan Wei}}
\def\confName{CVPR}
\def\confYear{2027}
\title{
    Reframe: Observation-Aligned Multi-Teacher\\
    On-Policy Distillation for Agentic Visual Reasoning\vspace{-12pt}}
\author{
    \textbf{Haozhi Yuan}$^{1,\dagger}$ \qquad
    \textbf{Handong Li}$^{2}$ \qquad
    \textbf{Yifan Zhang}$^{2}$ \qquad
    \textbf{Qianshan Wei}$^{2}$ \\[1.5mm]
    $^{1}$Nanjing University \qquad
    $^{2}$Institute of Automation, Chinese Academy of Sciences \\[1.5mm]
    {\small $^{\dagger}$Corresponding Author (\texttt{haozhiyuan@smail.nju.edu.cn})}
}

\begin{document}
\maketitle

\begin{abstract}
Agentic reinforcement learning (RL) and multi-teacher on-policy distillation (MOPD) offer complementary routes to post-training. For visual agents, however, tool use presents two related challenges. Models may overlook useful tool-returned observations, while poorly grounded actions can limit the evidence available for subsequent distillation. We introduce \textbf{Reframe}, a post-training framework that treats the observation state, rather than only the output distribution, as an object of alignment. First, \textbf{Ground-Drop-GRPO (GD-GRPO)} trains spatial and temporal specialists with evidence-aware rewards and stochastic removal of the original visual input, encouraging them to acquire grounded evidence and use the returned observations. Second, \textbf{Observation-Gated-MOPD (OG-MOPD)} replaces deficient tool actions with canonical ones mid-rollout and resumes distillation from the repaired observation. The two stages complement each other, as learning to rely on tool-returned observations can help the policy benefit from observation repair. Neither stage requires a task-specific SFT cold start. Extensive experiments over 14 image and video benchmarks demonstrate that Reframe outperforms both Mix-RL and vanilla MOPD. Ablation results show that GD-GRPO increases observation dependence and tool gain compared with outcome-based GRPO. \url{https://github.com/EliYuan30/Reframe}
\end{abstract}

\begin{figure}[t]
    \centering
    \includegraphics[width=\linewidth]{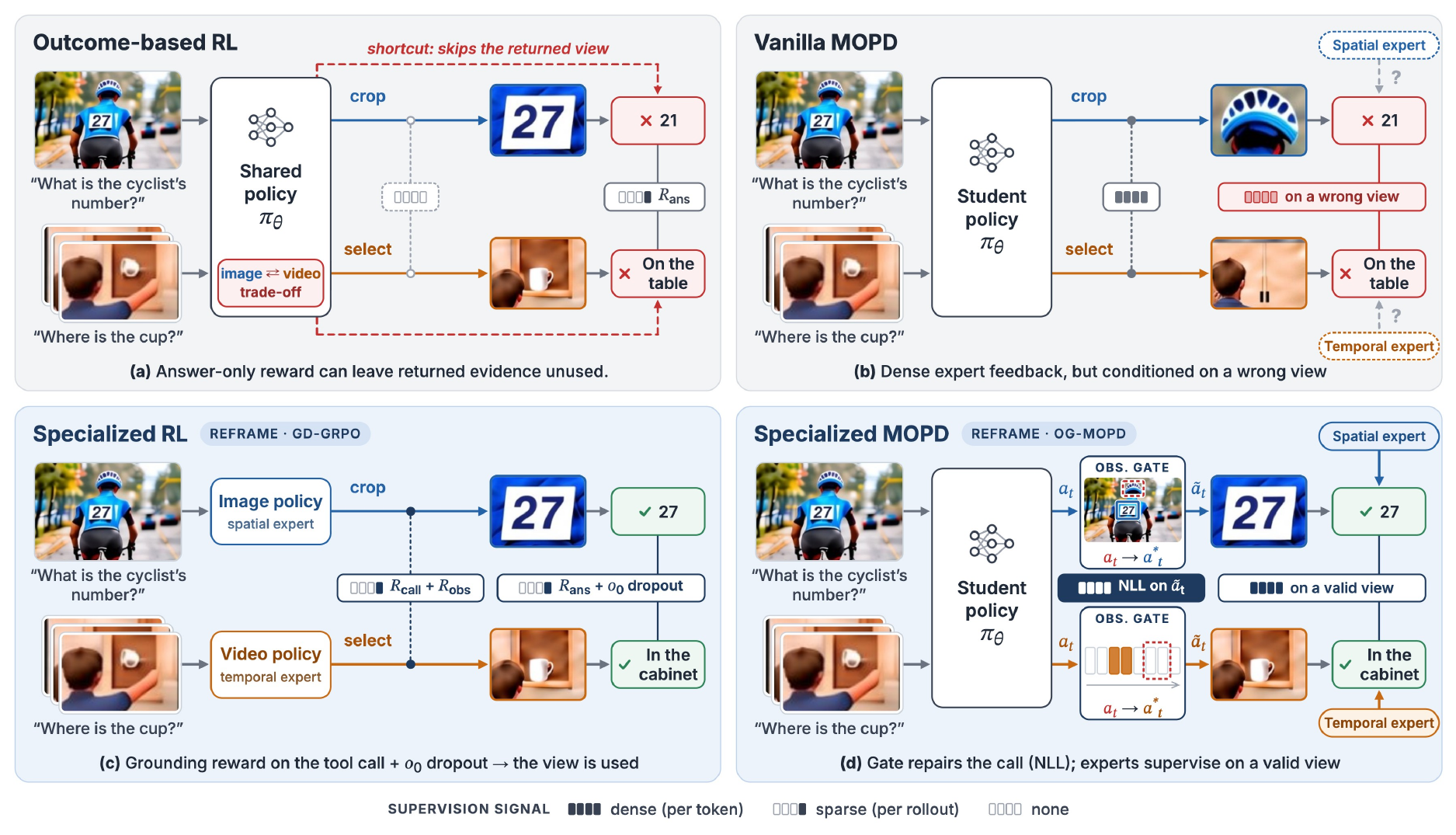}
    \vspace{-17pt}
    \caption{\textbf{Overview of Reframe: visual agents should produce reliable intermediate results and truly look at them.} (a)~Outcome-based RL without explicit grounding rewards can permit observation neglect. (b)~Vanilla MOPD may supervise trajectories conditioned on a poorly grounded observation state. (c)~Our GD-GRPO rewards grounded actions and encourages reliance on tool-returned observations. (d)~Our OG-MOPD repairs deficient tool actions before continuing distillation.}
    \label{fig:teaser}
    \vspace{-12pt}
\end{figure}

\section{Introduction}
\label{sec:introduction}

Agentic visual reasoning addresses the evidence-sparse bottleneck in perception-centered tasks by equipping vision-language models (VLMs) with tools to acquire fine-grained evidence. Related studies include \emph{think with images}~\citep{pixel-reasoner,thyme,deepeyes,deepeyesv2,adatooler-v,beacon,openai} and \emph{think with videos}~\citep{vital,video-mtr,framethinker,video-com}. However, invoking a tool does not guarantee that the model uses what it returns. We term this problem observation \emph{neglect}, where a model acquires an intermediate observation but its subsequent reasoning is largely unaffected~\citep{cauaudit}. Alongside efforts to improve when tools are invoked~\citep{adatooler-v,metis,beacon}, this raises a complementary question of how to train models to make effective use of the acquired observation.

Meanwhile, most previous works focus exclusively on a single modality, while the few that simultaneously incorporate image and video tools rely on a mixed-data SFT-then-RL training paradigm. However, this often leaves the final unified model significantly lagging behind the one trained on a single modality. We argue that this is the combined result of the pattern conflict of off-policy SFT training and the supervision sparsity of RL. A more promising alternative is a densely supervised, predominantly on-policy training paradigm. On-policy distillation (OPD) serves as a foundational approach: by supervising the student's own rollouts with dense token-level feedback, it bypasses both the sparse credit assignment of RL and the reference bias of off-policy imitation~\citep{gkd,opd-survey}. Extending this paradigm, multi-teacher on-policy distillation (MOPD)~\citep{mopd} routes student-generated trajectories to domain-specialized teachers, and has been increasingly adopted by frontier models~\citep{mimo-v2-flash,glm-52,kimi-k3,deepseek-v4,deepseek-v41-flash}. Subsequent works further refine it in terms of capability imbalance~\citep{open-mopd}, heterogeneous convergence~\citep{d3-mopd}, behavioral drift under tool use~\citep{toolcall-drift}, and cross-modal routing~\citep{ui-mopd,video-mopd}. However, existing paradigms primarily refine how supervision is delivered without intervening in rollout trajectories. This is a critical limitation for agentic visual reasoning since subsequent supervision is all conditioned on what tools returned.

\begin{table}[t]
\centering
\caption{\textbf{Comparison with representative visual agents.} Reframe adopts a predominantly on-policy paradigm to consolidate image and video reasoning within a unified model, while achieving the highest observation dependence and tool gain among the compared models. Observation dependence is measured on V$^*$. Tool gain is measured across seven image benchmarks.}
\label{tab:comparison}
\renewcommand{\arraystretch}{1.2}
\setlength{\tabcolsep}{3pt}
\footnotesize
\vspace{-4pt}
\begin{tabularx}{\columnwidth}{l
  >{\centering\arraybackslash\hsize=1.22\hsize}X
  >{\centering\arraybackslash\hsize=0.98\hsize}X
  >{\centering\arraybackslash\hsize=1.14\hsize}X
  >{\centering\arraybackslash\hsize=0.66\hsize}X}
\toprule
\multirow{3}{*}[-1.5pt]{\textbf{Model}}
& \multicolumn{2}{c}{\textit{Training Properties}}
& \multicolumn{2}{c}{\textit{Effective Tool Use}} \\
\cmidrule(lr){2-3}\cmidrule(l){4-5}
& \textbf{\makecell{Unified\\Image--Video}}
& \textbf{\makecell{No SFT\\Cold Start}}
& \textbf{\makecell{Observation\\Dependence}}
& \textbf{\makecell{Tool\\Gain}} \\
\midrule
Pixel-Reasoner & \cmark & \xmark & 23.25  & $+0.97$ \\
Thyme         & \xmark & \xmark & 34.20 & $+1.47$ \\
DeepEyes      & \xmark & \cmark & 19.47  & $+1.32$ \\
DeepEyesV2    & \xmark & \xmark & 41.93 & $-2.45$ \\
\midrule
\rowcolor{gray!10}
\textbf{Reframe\,\,(Ours)} & \cmark & \cmark & \textbf{75.53} & $\mathbf{+5.23}$ \\
\bottomrule
\vspace{-16pt}
\end{tabularx}
\end{table}

Motivated by these, we propose \textbf{Reframe}, a two-stage training framework comprising \textbf{Ground-Drop-GRPO (GD-GRPO)} and \textbf{Observation-Gated-MOPD (OG-MOPD)}. Together, the two training stages enforce the two core properties for effective tool use: GD-GRPO makes the acquired observation \emph{necessary}, while OG-MOPD keeps it \emph{valid}. GD-GRPO directly targets the \emph{neglect} problem by augmenting observation-based reward and stochastically dropping the original visual input. It forces the model to genuinely rely on the tool-returned results. Subsequently, OG-MOPD ensures that consolidation remains conditioned on valid trajectories by repairing deviated tool use. Across 14 image and video benchmarks, Reframe outperforms both Mix-RL and vanilla MOPD, even surpassing the dedicated experts on average. Furthermore, diagnostic analyses on observation dependence confirm that Reframe exhibits a markedly stronger causal reliance on tool-returned results than previous methods. 

Our contributions are summarized as follows:
\begin{itemize}
    \item We propose \textbf{Reframe}, the first tool-use MOPD framework that unifies image and video agentic reasoning. Its \textbf{OG-MOPD} actively intervenes in rollout trajectories and repairs deviated tool actions, promoting observation \emph{validity} throughout cross-modal consolidation.
    \item We address observation \emph{neglect}---the issue where agentic VLMs invoke a tool yet overlook what it returns. Through \textbf{GD-GRPO}, we enforce observation \emph{necessity} via input dropping and observation rewards.
    \item We develop the \textbf{Reframe-100K} dataset to support both training stages. Evaluations across 14 image and video benchmarks show consistently improved performance. Code, dataset, and model are open-sourced.
\end{itemize}

\section{Related Work}
\label{sec:related_work}

\paragraph{Agentic visual reasoning.}
Multimodal reasoning often uses textual chain-of-thought (CoT)~\citep{video-r1,onethinker}, but its benefits vary across visual tasks~\citep{vsi-bench,video-mme-v2,videoauto-r1}. Recent approaches instead incorporate visual information into reasoning~\citep{openai,latent-visual-reasoning}, including interface- or code-based tool calls that acquire query-relevant observations. These tools allow models to inspect image regions and video segments in finer detail that may be difficult to resolve from the original input. However, acquiring finer visual evidence does not guarantee its use. Recent studies find that tool-returned observations may have little causal effect on subsequent predictions~\citep{cauaudit,countercredit}.

\vspace{-8pt}
\paragraph{On-policy distillation.}
OPD provides dense token-level feedback on student-generated rollouts, reducing the state-distribution mismatch of off-policy imitation~\citep{gkd,opd-survey}. Effective transfer depends on compatible reasoning and novel teacher capabilities~\citep{rethink-opd-1}, coverage of student-visited states~\citep{rethink-opd-2}, and teacher corrections within the student's learnable token support~\citep{ta-opd}. Extensions support black-box teachers through logit-free verification~\citep{omni-opd}, temporal video grounding~\citep{video-opd}, and supervision through teacher-generated reasoning steps~\citep{step-opd}. These methods primarily refine supervision along student-generated trajectories. For visual agents, however, preceding tool actions also determine which observations are available when that supervision is applied. Token-level feedback on a continuation cannot retroactively recover evidence that the preceding action failed to acquire.

\vspace{-8pt}
\paragraph{Multi-teacher on-policy distillation.}
MOPD consolidates independently trained specialists by routing teacher supervision over a shared student's rollouts~\citep{mopd}, an approach increasingly used in frontier-model post-training~\citep{mimo-v2-flash,glm-52,kimi-k3,deepseek-v41-flash}. Open-MOPD studies imbalance from token budgets, convergence drift, and stale supervision~\citep{open-mopd}, while D$^3$-MOPD adapts domain sampling to different convergence rates~\citep{d3-mopd}. Multimodal variants route specialists for GUI interaction and video understanding~\citep{ui-mopd,video-mopd}. Other work identifies distortions of decision-critical tool tokens despite sufficient retained teacher probability mass~\citep{toolcall-drift}. These advances focus on routing and supervision while generally taking rollout observations as given. Reframe instead considers both whether acquired observations are used and whether they provide valid evidence for subsequent distillation, supporting unified image and video reasoning.

\section{Method}
\label{sec:method}

In this section, we introduce the data construction and the two-stage training procedure of Reframe. We make three design choices: (1)~We restrict the toolset to two fundamental vision tools, \emph{CropImage} and \emph{SelectFrames}, for simplicity. (2)~We adopt Qwen3.5-9B as our base model. (3)~Both RL and OPD stages use the same data corpus.

\subsection{Training Data Construction}
\label{sec:training_data_construction}

\begin{figure}[t]
    \centering
    \includegraphics[width=1\linewidth]{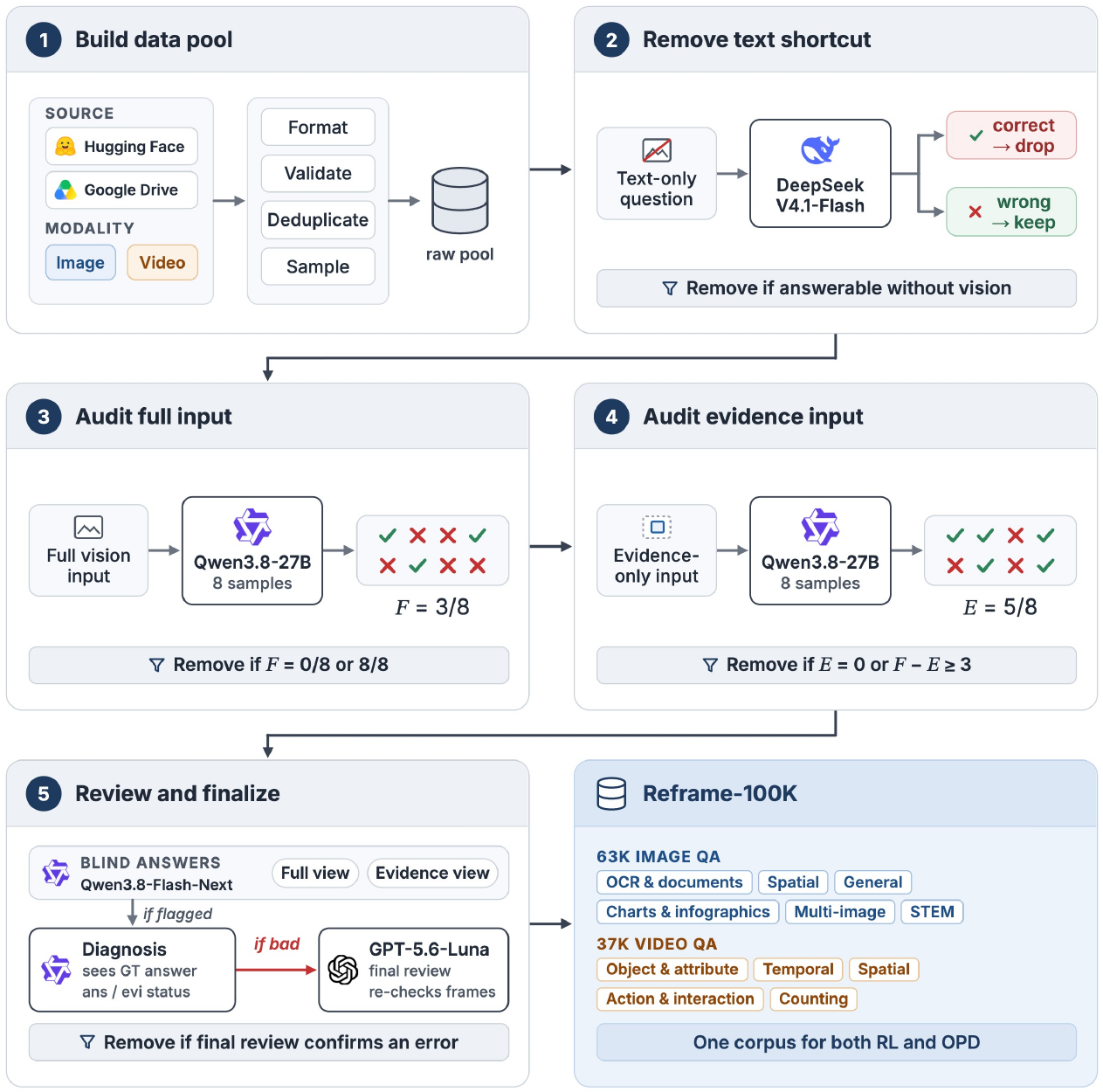}
    \caption{\textbf{Construction pipeline of Reframe-100K.} We first collect public data sources and preprocess them into a raw data pool. We then drop samples answerable from text alone, and audit the rest with 8 rollouts on the full input ($F$) and on the annotated evidence alone ($E$), discarding samples that are (1)~too easy or too hard; (2)~insufficiently supported by their evidence. Remaining suspicious samples are diagnosed by Qwen and removed only when a final reviewer confirms an answer or evidence error.}
    \label{fig:data_pipeline}
    \vspace{-12pt}
\end{figure}

Reframe-100K comprises 63K image QA pairs and 37K video QA pairs, constructed from heterogeneous spatial and temporal grounding annotations. As illustrated in Fig.~\ref{fig:data_pipeline}, we establish a five-step cascaded data construction pipeline guided by the following three principles.

\vspace{-8pt}
\paragraph{Quality control of grounding annotations.}
The quality of grounding annotations is critical to our framework, as unreliable evidence can undermine both evidence-aware rewards and subsequent observation repair. We therefore place emphasis on evidence quality throughout our data-cleaning pipeline, incorporating multiple rounds of evidence auditing and verification. As illustrated in Fig.~\ref{fig:data_pipeline}, these procedures aim to ensure that the retained annotations provide reliable supervision for visual tool use.

\vspace{-8pt}
\paragraph{Data-centric tool applicability.}
Determining whether a tool call is beneficial remains challenging, particularly when the model has not yet developed reliable tool-use capabilities. Unlike model-dependent strategies~\citep{adatooler-v,beacon}, we derive tool-applicability labels from annotation coverage, distinguishing beneficial, neutral, and non-beneficial examples without relying on base-model rollout accuracy or stronger-model judgments of tool utility. This avoids introducing model-prediction bias into applicability labels. Importantly, examples are not discarded solely because the base model can answer them without tools, as even correctly answered questions can provide valuable spatial or temporal tool-use supervision.

\vspace{-8pt}
\paragraph{Context-aware evidence.}
Grounding annotations typically localize minimal objects or temporal events, whereas visual question answering usually requires surrounding context for recognition, comparison, or disambiguation. We therefore expand each spatial bounding box by scaling its width and height by $1.5$ about its center, clipping the result to the image boundary. For temporal evidence, we extend each contiguous annotated segment by one frame at both ends. This context-preserving relaxation provides additional evidence without introducing excessive irrelevant visual content. We apply it only once before computing evidence coverage and constructing tool supervision.

Finally, spatial and temporal annotations are converted into executable tool metadata under a unified format. The resulting Reframe-100K supports both RL and OPD.

\subsection{GD-GRPO}
\label{sec:gd_grpo}

\begin{figure*}[t]
    \centering
    \includegraphics[width=\linewidth]{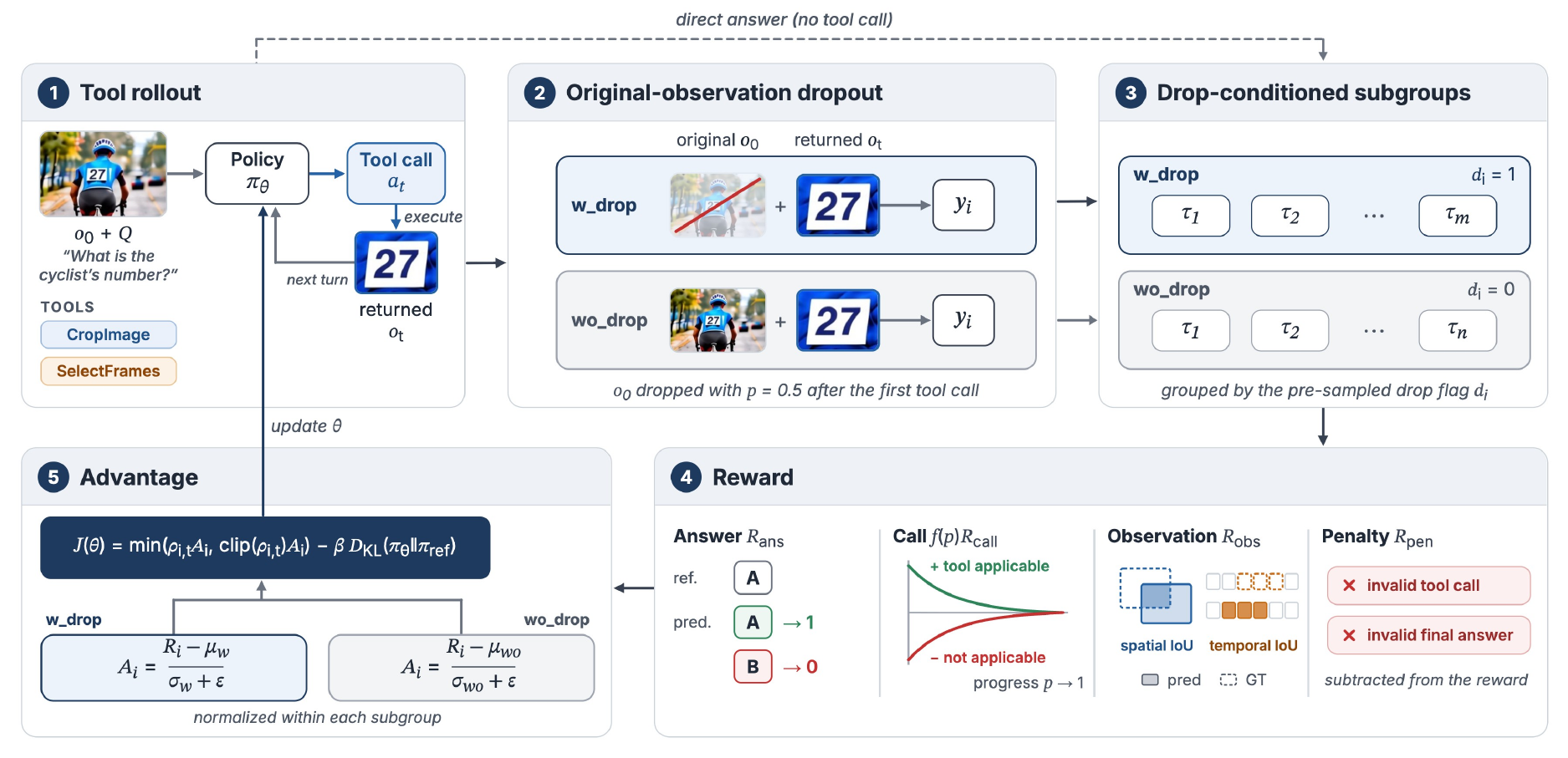}
    \vspace{-16pt}
    \caption{\textbf{Overview of GD-GRPO.} The policy samples multi-turn tool-use rollouts. After the first successful tool call, the original input $o_0$ is dropped with probability 0.5 while tool-returned observations $o_t$ remain visible, so the answer must rely on what the tools have retrieved. Rollouts are grouped by their pre-sampled drop flag $d_i$, and advantages are normalized within each subgroup so that dropout does not confound the comparison.}
    \label{fig:gd-grpo}
    \vspace{-8pt}
\end{figure*}

We train spatial and temporal specialists independently on the image and video subsets of Reframe-100K using GD-GRPO. The core idea is to couple \emph{observation acquisition} with \emph{observation reliance}. While observation rewards guide the policy toward informative tool actions, original-observation dropout reduces its tendency to bypass the resulting observations, thereby establishing the \emph{necessity} of the acquired observations. GD-GRPO is characterized by three key designs: reward, dropout, and subgroup.

\subsubsection{Reward Design}
Outcome correctness alone cannot stably reveal whether a tool retrieves relevant information or whether subsequent reasoning uses its returned observation. For an input $x=(Q,o_0)$ consisting of a question and its original observation, the reward is defined as
\begin{equation}
\label{eq:gd_reward}
\begin{aligned}
R(\tau;x,p)
&= R_{\mathrm{ans}}(\tau;x)
+ f(p)R_{\mathrm{call}}(\tau;x) \\
&\quad + \lambda_{\mathrm{obs}}R_{\mathrm{obs}}(\tau;x)
- R_{\mathrm{pen}}(\tau;x),
\end{aligned}
\end{equation}
where $\tau$ and $p\in[0,1]$ denote trajectories and training progress, respectively. The four terms evaluate answer correctness, tool call behavior, grounding quality, and output validity, where the tool call term provides a temporary exploration incentive derived from the annotation-based applicability. Both the valid-call bonus and the inapplicable-call penalty decay with training progress according to
\begin{equation}
\label{eq:call_decay}
f(p)
=
\frac{\exp(-\kappa p)-\exp(-\kappa)}
     {1-\exp(-\kappa)},
\qquad \kappa>0.
\end{equation}
Unlike $R_{\mathrm{call}}$, $R_{\mathrm{obs}}$ remains active throughout training. For images, we jointly match the cropped images along a trajectory to the annotated regions using global one-to-one IoU matching, which rewards complementary coverage while preventing repeated localization of the same target from multiplying its reward. For videos, we evaluate the latest attempt using frame IoU rather than retrospectively selecting the most favorable attempt. Later attempts are discounted to encourage efficient acquisition. Importantly, grounding supervision is not gated by answer correctness. A proper tool action receives credit even when the policy does not answer correctly. Examples without grounding annotations receive no $R_{\mathrm{call}}$ or $R_{\mathrm{obs}}$, and remain supervised by answer correctness and validity penalties.

\subsubsection{Original-Observation Dropout}
Accurate localization does not by itself establish observation reliance. A model may still answer from the original visual input without using the tool-returned result. To reduce this bypass, we stochastically drop $o_0$ with a probability of $0.5$ after the first successful tool call during training, while retaining the question, textual history, and tool-returned observations. The model then has to rely on the acquired observations for subsequent reasoning rather than bypassing them through the original input. At inference, the original visual input is always retained.

Rather than prescribing tool actions, the dropout changes the conditions under which they contribute to successful reasoning. By removing the original visual input, it turns tool-returned observations from supplementary cues into essential information for subsequent reasoning. Coupled with grounding rewards, it encourages the policy to both acquire informative observations and genuinely rely on them.

\subsubsection{Drop-Conditioned Subgroup}
Original-observation dropout introduces an external source of variation into rollout difficulty. Comparing dropped and retained trajectories against a shared group baseline may encourage lower-quality trajectories, thereby confounding action quality with the randomly assigned condition. To address this, we compute relative advantages within subgroups that share the same pre-sampled dropout indicator $d_i \in \{0, 1\}$. For rollouts $\{\tau_i\}$ of input $x$, define $\mathcal{G}_{x,d}=\{i:d_i=d\}$. The advantage is
\begin{equation}
\label{eq:drop_conditioned_advantage}
A_i
=
\begin{cases}
\displaystyle
\frac{R_i-\mu_{x,d_i}}
     {\sigma_{x,d_i}+\epsilon},
& |\mathcal{G}_{x,d_i}|>1,\\[6pt]
0,
& \text{otherwise}.
\end{cases}
\end{equation}
Crucially, subgroups are defined by the pre-sampled dropout indicator rather than actual dropout, ensuring that group assignment remains independent of the policy's tool-use decisions.

Overall, GD-GRPO couples observation acquisition with downstream reliance: grounding rewards encourage the policy to acquire informative observations, while original-vision dropout promotes their actual use. This provides the foundation for observation-aligned capability consolidation in the subsequent OG-MOPD stage.

\subsection{OG-MOPD}
\label{sec:og_mopd}

For tool-use models, the effectiveness of distillation depends not only on the selected teacher but also on the observations available when its supervision is applied. A deficient tool action can place the student in an uninformative visual state, constraining all subsequent teacher feedback. Although token-level optimization can improve future actions, feedback on the current continuation does not retroactively repair the observation already returned. OG-MOPD therefore intervenes at the tool-action boundary to improve observation \emph{validity}, while retaining the student's generation. Essentially, this is a trade-off between on-policy learning and observation validity.

\begin{figure}[t]
    \centering
    \includegraphics[width=\linewidth]{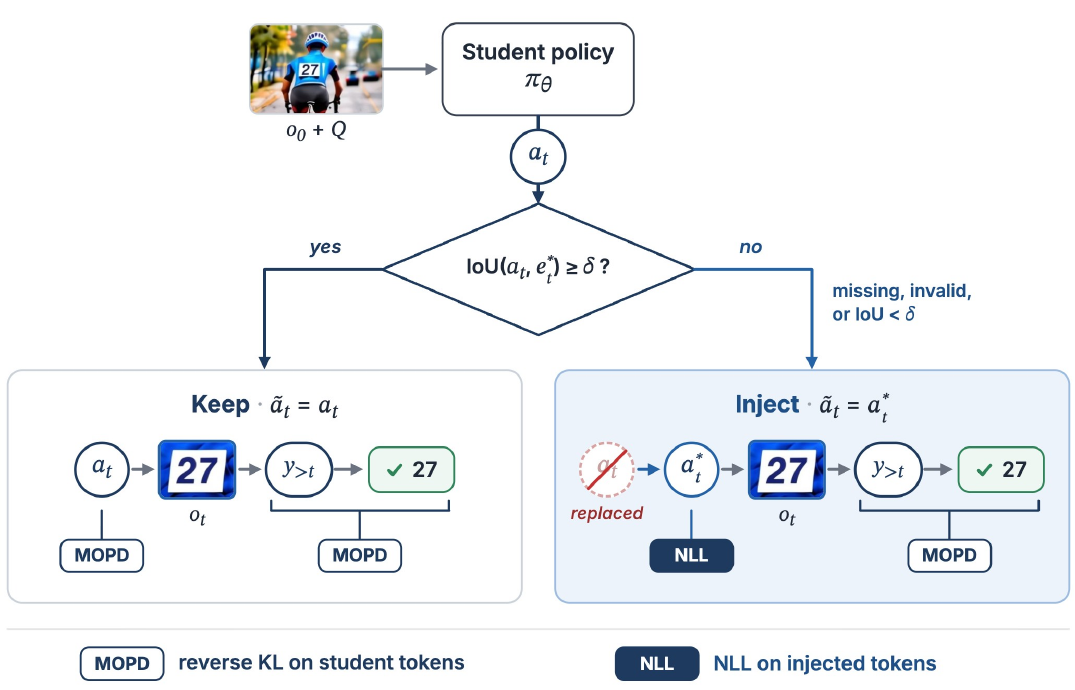}
    \vspace{-16pt}
    \caption{\textbf{Overview of OG-MOPD.} Tool actions of the student policy are checked against annotated evidence using an IoU threshold. Missing, invalid, or poorly grounded actions are replaced with annotation-derived actions to obtain query-relevant observations. Injected action tokens receive NLL supervision, while retained student-generated tokens are trained with MOPD.}
    \label{fig:og-mopd}
    \vspace{-8pt}
\end{figure}

\subsubsection{Observation-Gated Rollouts}
Let $a_t$ denote the student's proposed tool action, $a_t^\star$ the canonical action derived from grounding annotations, and $e_t^\star$ the corresponding annotated evidence. For tool-applicable examples, we selectively replace missing, invalid, or poorly grounded actions according to
\begin{equation}
\label{eq:observation_gate}
\tilde a_t =
\begin{cases}
a_t^\star, & a_t \text{ is invalid or }
             \operatorname{IoU}(a_t,e_t^\star)<\delta,\\
a_t, & \text{otherwise},
\end{cases}
\end{equation}
where $\delta$ is the grounding threshold. Missing actions are treated as invalid, and IoU is measured against the original or context-expanded annotations.

The environment executes $\tilde a_t$ and returns the corresponding observation, from which the student resumes sampling. Crucially, this selective intervention preserves well-grounded student actions while repairing deficient ones, balancing on-policy exploration with observation validity.

\subsubsection{Branch-Aware Distillation}

Observation repair introduces two distinct token sources: student-sampled tokens $\mathcal{S}$ and injected action tokens $\mathcal{I}$. For the former, we retain standard MOPD supervision, where the corresponding image or video specialist provides dense token-level feedback on the student's actual continuation. Teacher routing follows the executed tool action, with the input modality providing the initial assignment. However, injected actions are not sampled from the student policy and should not be treated as on-policy tokens. We therefore introduce a separate NLL objective:
\begin{equation}
\label{eq:og_mopd_objective}
\begin{aligned}
\mathcal{L}_{\mathrm{OG\text{-}MOPD}}
&=
\mathcal{L}_{\mathrm{MOPD}}(\mathcal{S})
+
\lambda_{\mathrm{br}}\mathcal{L}_{\mathrm{NLL}}(\mathcal{I}),\\
\mathcal{L}_{\mathrm{NLL}}(\mathcal{I})
&=
-\frac{1}{|\mathcal{I}|}
\sum_{k\in\mathcal{I}}
\log\pi_\theta(y_k\mid h_k),
\end{aligned}
\end{equation}
where $h_k$ denotes the interaction history preceding token $y_k$. If there is no injected action, then the NLL term is omitted. This separation provides complementary supervision: NLL teaches the student how to reach informative observation states, while MOPD transfers expert behavior along the subsequent student-sampled trajectories. At inference, neither observation gating nor specialist teachers are required. Training details are provided in Appendix~\ref{app:training_details}.

\section{Experiments}

\subsection{Experimental Setup}

\paragraph{Benchmarks.}
We evaluate Reframe on seven image benchmarks and seven video benchmarks, testing whether targeted visual inspection translates into both fine-grained perception and broader visual understanding. For images, V$^*$~\citep{V*}, HR-Bench 4K/8K~\citep{hrbench}, VisualProbe~\citep{visualprobe}, and ZoomBench~\citep{zoombench} stress the ability to find and resolve task-relevant details within complex visual inputs. MME-RealWorld~\citep{mme-realworld} extends the evaluation to diverse real-world scenes, while PerceptionBench~\citep{perceptionbench} probes atomic visual perception. For videos, Video-MME~\citep{video-mme}, Video-MME-v2~\citep{video-mme-v2}, MVBench~\citep{mvbench}, and MLVU~\citep{mlvu} assess understanding across diverse tasks and temporal scales. MME-VideoOCR~\citep{mme-videoocr} and VSI-Bench~\citep{vsi-bench} target video text understanding and spatial reasoning, respectively. V-STaR~\citep{vstar} probes spatial and temporal evidence understanding.

\vspace{-8pt}
\paragraph{Baselines.}
We conduct two groups of comparisons. The first compares training paradigms using five baselines: Qwen3.5-9B; Image-RL and Video-RL, the modality-specific specialists trained with GD-GRPO on single-modality data; Mix-RL, trained with GD-GRPO on the combined image and video data; and MOPD, which consolidates the two specialists through vanilla MOPD. The second compares Reframe with existing agentic visual reasoning models: Pixel-Reasoner, Thyme, DeepEyes, and DeepEyesV2, in terms of performance, observation dependence, and tool gain. Furthermore, outcome-based GRPO serves as a baseline to diagnose the proposed GD-GRPO.

\vspace{-12pt}
\paragraph{Evaluation protocol.}
We evaluate all models under shared visual-token budgets using greedy decoding. Each original image receives up to 5,120 tokens, and each tool-returned image receives up to 1,960 tokens. For videos, we use 64 frames with 128 tokens per frame. Each selected frame receives up to 400 tokens, subject to a cumulative budget of 4,096 tokens across all select calls. Evaluation details are provided in Appendix~\ref{app:evaluation_protocol}.

\subsection{Main Results}

\begin{table*}[t]
\centering
\caption{\textbf{Comparison of training paradigms.} Bold and underline mark the best and second-best scores. Image-RL, Video-RL, and Mix-RL use GD-GRPO; MOPD and Reframe distill the same specialists with vanilla MOPD and OG-MOPD. Video-MME-v2, VSI-Bench, and V-STaR report Non-Lin Score, accuracy/MRA, and $100\times\mathrm{mAM}$, respectively; MLVU reports M-Avg.}
\label{tab:training_paradigms}
\renewcommand{\arraystretch}{1.15}
\setlength{\tabcolsep}{4pt}
\small
\vspace{-6pt}
\begin{tabular*}{\textwidth}{@{\extracolsep{\fill}}lcccccc@{}}
\toprule
\rowcolor{white}
\textbf{Benchmark} & \textbf{Qwen3.5-9B} & \textbf{Image-RL} & \textbf{Video-RL} & \textbf{Mix-RL} & \textbf{MOPD} & \textbf{Reframe (Ours)} \\
\midrule
\rowcolor{imagegold!10}
\multicolumn{7}{c}{\textit{Image Understanding}} \\
V$^*$ & 83.77 & 89.53 & 86.91 & 88.48 & \underline{90.05} & \imagescoregain{\textbf{91.62}}{+7.85} \\
HR-Bench-4K & 83.13 & \textbf{85.25} & 82.63 & 83.25 & \underline{84.88} & \imagescoregain{84.75}{+1.62} \\
HR-Bench-8K & 76.88 & \underline{82.63} & 81.63 & 80.50 & 82.50 & \imagescoregain{\textbf{83.50}}{+6.62} \\
VisualProbe & 43.88 & 51.65 & 51.65 & 51.46 & \underline{52.62} & \imagescoregain{\textbf{57.48}}{+13.60} \\
ZoomBench & 50.53 & 57.40 & 56.57 & 57.51 & \underline{58.46} & \imagescoregain{\textbf{59.88}}{+9.35} \\
MME-RealWorld & 71.45 & 72.77 & 71.12 & 70.99 & \underline{72.99} & \imagescoregain{\textbf{73.04}}{+1.59} \\
PerceptionBench & 34.60 & \textbf{36.23} & 33.97 & 35.07 & 35.80 & \imagescoregain{\underline{36.07}}{+1.47} \\
\cmidrule(lr){1-7}
\textbf{Image Avg.} & 63.46 & 67.92 & 66.35 & 66.75 & \underline{68.19} & \imagescoregain{\textbf{69.48}}{+6.02} \\
\midrule
\rowcolor{videoblue!10}
\multicolumn{7}{c}{\textit{Video Understanding}} \\
Video-MME & 65.00 & 66.30 & \underline{67.93} & 66.44 & 67.41 & \videoscoregain{\textbf{69.07}}{+4.07} \\
Video-MME-v2 & 14.12 & 13.55 & \underline{14.47} & 13.46 & 14.24 & \videoscoregain{\textbf{15.61}}{+1.49} \\
MVBench & 66.08 & 68.50 & \textbf{70.37} & 69.29 & \underline{70.08} & \videoscoregain{69.11}{+3.03} \\
MLVU & 63.83 & 66.42 & \textbf{70.39} & 68.54 & 69.48 & \videoscoregain{\underline{70.10}}{+6.27} \\
MME-VideoOCR & 62.15 & 64.26 & 63.74 & 63.49 & \underline{64.92} & \videoscoregain{\textbf{65.28}}{+3.13} \\
VSI-Bench & 52.97 & 50.40 & \textbf{56.74} & 53.45 & 55.15 & \videoscoregain{\underline{56.70}}{+3.73} \\
V-STaR & 24.09 & 23.29 & \textbf{25.82} & 25.19 & 24.98 & \videoscoregain{\underline{25.80}}{+1.71} \\
\cmidrule(lr){1-7}
\textbf{Video Avg.} & 49.75 & 50.39 & \underline{52.78} & 51.41 & 52.32 & \videoscoregain{\textbf{53.10}}{+3.35} \\
\bottomrule
\end{tabular*}
\vspace{-6pt}
\end{table*}


\begin{table}[t]
\centering
\caption{\textbf{GD-GRPO versus outcome-based GRPO for image-specialist training.} Image performance and tool gain average seven benchmarks. On V$^*$, observation dependence is $100\times\frac{\mathrm{GT}-(\mathrm{Random}+\mathrm{Blank})/2}{100-(\mathrm{Random}+\mathrm{Blank})/2}$ (\%).}
\label{tab:gd_grpo_ablation}
\renewcommand{\arraystretch}{1.15}
\setlength{\tabcolsep}{8pt}
\small
\vspace{-6pt}
\begin{tabular*}{\columnwidth}{@{\extracolsep{\fill}}lcc@{}}
\toprule
\textbf{Condition} & \textbf{GRPO} & \textbf{GD-GRPO} \\
\midrule
\rowcolor{imagegold!10}
\multicolumn{3}{c}{\textit{Image Understanding}} \\
\textbf{Image Avg.} & 65.85 & 67.92 \\
\midrule
\rowcolor{dependencepurple!10}
\multicolumn{3}{c}{\textit{Observation Dependence}} \\
Natural & 86.39 & 89.53 \\
Direct & 85.34 & 87.43 \\
GT & 87.96 & 93.19 \\
Random & 72.25 & 75.92 \\
Blank & 76.96 & 80.10 \\
\cmidrule(lr){1-3}
\textbf{Dependence} & 52.59 & 69.03 \\
\midrule
\rowcolor{toolteal!10}
\multicolumn{3}{c}{\textit{Tool Gain}} \\
With Tools & 65.85 & 67.92 \\
Without Tools & 63.88 & 64.10 \\
\cmidrule(lr){1-3}
\textbf{Gain} & \gain{+1.97} & \gain{+3.82} \\
\bottomrule
\end{tabular*}
\vspace{-8pt}
\end{table}


\begin{table*}[t]
\centering
\caption{\textbf{Comparison with existing visual agents.} Bold and underline mark the best and second-best averages. Video metrics follow Table~\ref{tab:training_paradigms}. On V$^*$, observation dependence is $100\times\frac{\mathrm{GT}-(\mathrm{Random}+\mathrm{Blank})/2}{100-(\mathrm{Random}+\mathrm{Blank})/2}$ (\%). Tool gain is averaged over seven image benchmarks.}
\label{tab:existing_models}
\renewcommand{\arraystretch}{1.15}
\setlength{\tabcolsep}{4pt}
\small
\vspace{-4pt}
\begin{tabular*}{\textwidth}{@{\extracolsep{\fill}}lccccc@{}}
\toprule
\rowcolor{white}
\textbf{Benchmark/Condition} & \textbf{Pixel-Reasoner} & \textbf{Thyme} & \textbf{DeepEyes} & \textbf{DeepEyesV2} & \textbf{Reframe (Ours)} \\
\midrule
\rowcolor{imagegold!10}
\multicolumn{6}{c}{\textit{Image Understanding}} \\
V$^*$ & 81.68 & 81.68 & 83.25 & 81.68 & 91.62 \\
HR-Bench-4K & 72.50 & 77.13 & 74.63 & 71.00 & 84.75 \\
HR-Bench-8K & 65.88 & 67.88 & 67.50 & 61.38 & 83.50 \\
VisualProbe & 34.95 & 41.75 & 37.67 & 33.20 & 57.48 \\
ZoomBench & 40.47 & 46.51 & 46.75 & 41.30 & 59.88 \\
MME-RealWorld & 63.12 & 65.26 & 64.35 & 64.12 & 73.04 \\
PerceptionBench & 24.30 & 24.20 & 25.87 & 25.70 & 36.07 \\
\cmidrule(lr){1-6}
\textbf{Image Avg.} & 54.70 & \underline{57.77} & 57.15 & 54.05 & \textbf{69.48} \\
\midrule
\rowcolor{videoblue!10}
\multicolumn{6}{c}{\textit{Video Understanding}} \\
Video-MME & 61.11 & 60.93 & 58.74 & 59.56 & 69.07 \\
Video-MME-v2 & 10.84 & 11.01 & 10.32 & 10.61 & 15.61 \\
MVBench & 62.82 & 61.92 & 59.16 & 59.63 & 69.11 \\
MLVU & 64.64 & 64.45 & 61.97 & 63.85 & 70.10 \\
MME-VideoOCR & 54.62 & 53.79 & 54.10 & 55.08 & 65.28 \\
VSI-Bench & 33.22 & 34.93 & 32.36 & 33.44 & 56.70 \\
V-STaR & 20.39 & 22.65 & 19.10 & 22.42 & 25.80 \\
\cmidrule(lr){1-6}
\textbf{Video Avg.} & 43.95 & \underline{44.24} & 42.25 & 43.51 & \textbf{53.10} \\
\midrule
\rowcolor{dependencepurple!10}
\multicolumn{6}{c}{\textit{Observation Dependence}} \\
Natural & 81.68 & 81.68 & 83.25 & 81.68 & 91.62 \\
Direct & 75.39 & 79.06 & 81.15 & 83.77 & 87.43 \\
GT & 82.72 & 86.91 & 84.82 & 85.86 & 96.86 \\
Random & 76.44 & 79.06 & 81.15 & 72.77 & 85.86 \\
Blank & 78.53 & 81.15 & 81.15 & 78.53 & 88.48 \\
\cmidrule(lr){1-6}
\textbf{Dependence} & 23.25 & 34.20 & 19.47 & \underline{41.93} & \textbf{75.53} \\
\midrule
\rowcolor{toolteal!10}
\multicolumn{6}{c}{\textit{Tool Gain}} \\
With Tools & 54.70 & 57.77 & 57.15 & 54.05 & 69.48 \\
Without Tools & 53.73 & 56.30 & 55.83 & 56.50 & 64.25 \\
\cmidrule(lr){1-6}
\textbf{Gain} & \gain{+0.97} & \gain{\underline{+1.47}} & \gain{+1.32} & \regression{-2.45} & \gain{\textbf{+5.23}} \\
\bottomrule
\end{tabular*}
\vspace{-4pt}
\end{table*}

Results are shown in Tables~\ref{tab:training_paradigms}, \ref{tab:gd_grpo_ablation} and \ref{tab:existing_models}. We make three main observations: (1)~Reframe outperforms vanilla MOPD, which in turn outperforms Mix-RL. MOPD can even surpass its specialist teachers, while observation repair adds further gains; (2)~Compared with outcome-based GRPO, GD-GRPO improves the specialist's benchmark performance and increases observation dependence and tool gain; and (3)~Reframe achieves the largest tool gain and the strongest dependence on tool-returned observations among the compared visual agents.

\subsection{Ablation and Analysis}

We first examine GD-GRPO, then compare specialist consolidation methods, and finally analyze observation dependence and tool gain against existing agents.

\subsubsection{Effectiveness of GD-GRPO}

Table~\ref{tab:gd_grpo_ablation} compares image specialists trained with GD-GRPO and outcome-based GRPO, which uses only correctness rewards and no original-observation dropout. GD-GRPO improves the image average by 2.07 points. In the tool-gain evaluation, enabling tools adds 3.82 points with GD-GRPO versus 1.97 with GRPO, while tool-free accuracy improves by only 0.22 points. These results suggest that GD-GRPO yields only modest gains in general visual capabilities, while its main benefit lies in enabling the model to translate visual tool use into better reasoning performance.

To better understand this tool-related improvement, we examine how GD-GRPO changes the model's response to different tool observations. GD-GRPO improves accuracy under GT, Random, and Blank by 5.23, 3.67, and 3.14 points, respectively, with the largest gain under GT. We summarize this behavior with the dependence score (Appendix~\ref{app:observation_dependence}), defined as the fraction of errors under uninformative observations (the mean of Random and Blank) that are corrected once the ground-truth observation is supplied. It measures sensitivity to observation content rather than overall capability. GD-GRPO raises it from 52.59 to 69.03, while also improving accuracy under corrupted observations. Together with the larger tool gain, these results support better use of returned observations rather than greater vulnerability to misleading observations.

\subsubsection{Comparison of Training Paradigms}

We compare Mix-RL, MOPD, and Reframe against the modality-specific experts in Table~\ref{tab:training_paradigms}.

\vspace{-8pt}
\paragraph{Mix-RL falls short of the specialists.}
Mix-RL trails Image-RL and Video-RL in their respective modalities by 1.17 and 1.37 points (Table~\ref{tab:training_paradigms}), retaining 74\% and 55\% of their gains over the base model. Although it uses the GD-GRPO reward, these signals are aggregated into trajectory-level advantages rather than separate assessments of each acquisition and reasoning decision. The shared policy must therefore discover spatial and temporal strategies jointly, yet gradient updates favoring one modality do not necessarily benefit the other. This couples exploration across modalities with coarse credit assignment, providing a plausible explanation for the specialist gap. In contrast, the MOPD results show that near-specialist performance in both modalities is attainable, suggesting that the learning process does matter.

\vspace{-8pt}
\paragraph{MOPD approaches specialist performance.}
MOPD retains 106\% and 85\% of the gains achieved by Image-RL and Video-RL, respectively. It surpasses both teachers on 5 of the 14 benchmarks. The key difference from Mix-RL lies in the supervision density. Mix-RL uses a trajectory-level advantage, while MOPD provides specialist feedback for each generated token. This gives the student more direct guidance on intermediate generation choices, reducing the need to infer their value from an aggregate response-level signal. Such dense supervision can make better use of each rollout, making MOPD a more practical choice for consolidating specialists.

\vspace{-8pt}
\paragraph{Observation repair yields additional gains.}
With the same GD-GRPO specialists as teachers, Reframe improves over MOPD by 1.29 points on images and 0.78 on videos. The largest gain occurs on VisualProbe (+4.86). MOPD already teaches both actions and reasoning, but its feedback on a continuation is conditioned on the observation the student has obtained. Repair provides task-relevant evidence before distillation continues. Its contribution is therefore focused on deficient actions in tool-applicable examples, complementing the broader consolidation already achieved by MOPD. The central trade-off in observation repair is between observation validity and the degree to which the rollout remains on-policy. An externally supplied action changes the history from which the student continues, even though the continuation itself remains student-generated. OG-MOPD limits this intervention by preserving valid student actions and repairing only deficient ones. NLL on the injected actions encourages the student to produce the corrected actions itself at inference.

\vspace{-8pt}
\paragraph{Visual reasoning partially transfers across modalities.}
Image-RL improves the video average from 49.75 to 50.39, while Video-RL raises the image average from 63.46 to 66.35. The transferable component may include primitive visual capabilities such as entity recognition, OCR, and spatial intelligence, as well as a shared tendency to seek fine-grained evidence. Grounding-oriented training can improve general perception beyond its training task.

\subsubsection{Comparison with Existing Agentic Models}

Reframe leads on all 14 benchmarks in Table~\ref{tab:existing_models}, exceeding the strongest external image and video averages by 11.71 and 8.86 points, respectively. Under shared visual budgets, we examine how models use acquired observations and benefit from tools.

\vspace{-8pt}
\paragraph{Reframe uses the evidence it acquires.}
On V$^*$, we examine model performance with ground-truth (GT), random, and blank cropped images as tool-returned observations. Direct evaluates the model without tools, while Natural lets the model decide when and how to use them. Reframe's dependence score is 75.53, compared with at most 41.93 for the other models. Its accuracy reaches 96.86 with GT, versus 85.86 with Random and 88.48 with Blank, despite the original image remaining available. Reliance on returned observations can be beneficial when tool actions reliably acquire relevant evidence. Reframe's larger gap across conditions, strong GT performance, and positive tool gain together suggest that it can translate this reliance into improved reasoning performance.

\vspace{-8pt}
\paragraph{Tools contribute more to Reframe.}
Tool gain complements the controlled-observation analysis by measuring the benefit of autonomous tool use. Across seven image benchmarks, tools improve Reframe from 64.25 to 69.48, a gain of 5.23 points versus at most 1.47 for the other models. Relative to the tool-free Qwen3.5-9B (63.46), 87\% of the overall gain stems from tool use, indicating that Reframe's advantage largely arises from interaction with visual tools. Useful observations must be acquired through the model's own actions. The larger tool gain, therefore, provides evidence that Reframe can combine evidence acquisition and use to improve its final answers.

\section{Conclusion}

\vspace{-4pt}
Reframe consolidates image and video reasoning predominantly through on-policy training. GD-GRPO encourages reliance on tool-returned observations through grounding rewards and original-input dropout, while OG-MOPD repairs deficient tool actions before continuing distillation. Across 14 benchmarks, Reframe improves over Mix-RL and vanilla MOPD, even surpassing both specialists. Diagnostic evaluations show stronger observation dependence and larger tool gains than the compared agentic visual models, highlighting the importance of both acquiring useful evidence and learning to use it.

\clearpage
{\small
\bibliographystyle{ieeenat_fullname}
\bibliography{main}

@inproceedings{V*,
  author    = {Penghao Wu and Saining Xie},
  title     = {{V$^*$}: Guided Visual Search as a Core Mechanism in Multimodal {LLMs}},
  booktitle = {CVPR},
  year      = {2024}
}

@inproceedings{hrbench,
  author    = {Wenbin Wang and Liang Ding and Minyan Zeng and Xiabin Zhou and Li Shen and Yong Luo and Wei Yu and Dacheng Tao},
  title     = {Divide, Conquer and Combine: A Training-Free Framework for High-Resolution Image Perception in Multimodal Large Language Models},
  booktitle = {AAAI},
  year      = {2025}
}

@inproceedings{visualprobe,
  author    = {Xin Lai and Junyi Li and Wei Li and Tao Liu and Tianjian Li and Hengshuang Zhao},
  title     = {{Mini-o3}: Scaling Up Reasoning Patterns and Interaction Turns for Visual Search},
  booktitle = {ICLR},
  year      = {2026}
}

@inproceedings{zoombench,
  author    = {Lai Wei and Liangbo He and Jun Lan and Lingzhong Dong and Yutong Cai and Siyuan Li and Huijia Zhu and Weiqiang Wang and Linghe Kong and Yue Wang and Zhuosheng Zhang and Weiran Huang},
  title     = {Zooming without Zooming: Region-to-Image Distillation for Fine-Grained Multimodal Perception},
  booktitle = {ICML},
  year      = {2026}
}

@inproceedings{mme-realworld,
  author    = {Yi-Fan Zhang and Huanyu Zhang and Haochen Tian and Chaoyou Fu and Shuangqing Zhang and Junfei Wu and Feng Li and Kun Wang and Qingsong Wen and Zhang Zhang and Liang Wang and Rong Jin and Tieniu Tan},
  title     = {{MME-RealWorld}: Could Your Multimodal {LLM} Challenge High-Resolution Real-World Scenarios that are Difficult for Humans?},
  booktitle = {ICLR},
  year      = {2025}
}

@article{perceptionbench,
  author        = {Zichao Lin and Yifeng Xie and Bowen Qu and Haiming Wang and Jia Li and Haoning Wu and Yuhao Dong and Zuhao Yang and Jinguo Zhu and Haoyu Lu and Zijia Zhao and Tongtian Yue and Zhangyang Qi and Junwei Yang and Mengfan Dong and Peizhou Cao and Chenzhuang Du and Zaida Zhou and Haotian Yao and Hao Yang and Hongcheng Gao and Lin Sui and Weihong Li and Xinxing Zu and Jia Chen and Yao Wang and Xiaoxue Wu and Yalin Wang and Y. Charles and Yiping Bao and Yangyang Liu and Zhiqi Huang and Xinyu Zhou},
  title         = {{PerceptionBench}: Evaluating Atomic Visual Perception in Multimodal Large Language Models},
  journal       = {arXiv:2607.24957},
  year          = {2026},
  eprint        = {2607.24957},
  archivePrefix = {arXiv},
  primaryClass  = {cs.CV}
}

@inproceedings{vstar,
  author    = {Zixu Cheng and Jian Hu and Ziquan Liu and Chenyang Si and Wei Li and Shaogang Gong},
  title     = {{V-STaR}: Benchmarking {Video-LLMs} on Video Spatio-Temporal Reasoning},
  booktitle = {CVPR Findings},
  year      = {2026}
}

@inproceedings{vsi-bench,
  author    = {Jihan Yang and Shusheng Yang and Anjali W. Gupta and Rilyn Han and Fei-Fei, Li and Saining Xie},
  title     = {Thinking in Space: How Multimodal Large Language Models See, Remember, and Recall Spaces},
  booktitle = {CVPR},
  year      = {2025}
}

@article{video-mme-v2,
  author  = {Chaoyou Fu and Haozhi Yuan and Yuhao Dong and Yi-Fan Zhang and Yunhang Shen and Xiaoxing Hu and Xueying Li and Jinsen Su and Chengwu Long and Xiaoyao Xie and Yongkang Xie and Xiawu Zheng and Xue Yang and Haoyu Cao and Yunsheng Wu and Ziwei Liu and Xing Sun and Caifeng Shan and Ran He},
  title   = {{Video-MME-v2}: Towards the Next Stage in Benchmarks for Comprehensive Video Understanding},
  journal = {arXiv:2604.05015},
  year    = {2026}
}

@inproceedings{video-mme,
  author    = {Chaoyou Fu and Yuhan Dai and Yongdong Luo and Lei Li and Shuhuai Ren and Renrui Zhang and Zihan Wang and Chenyu Zhou and Yunhang Shen and Mengdan Zhang and Peixian Chen and Yanwei Li and Shaohui Lin and Sirui Zhao and Ke Li and Tong Xu and Xiawu Zheng and Enhong Chen and Caifeng Shan and Ran He and Xing Sun},
  title     = {{Video-MME}: The First-Ever Comprehensive Evaluation Benchmark of Multi-modal {LLMs} in Video Analysis},
  booktitle = {CVPR},
  year      = {2025}
}

@inproceedings{mlvu,
  author    = {Junjie Zhou and Yan Shu and Bo Zhao and Boya Wu and Zhengyang Liang and Shitao Xiao and Minghao Qin and Xi Yang and Yongping Xiong and Bo Zhang and Tiejun Huang and Zheng Liu},
  title     = {{MLVU}: Benchmarking Multi-task Long Video Understanding},
  booktitle = {CVPR},
  year      = {2025}
}

@inproceedings{mme-videoocr,
  author    = {Yang Shi and Huanqian Wang and Wulin Xie and Huanyao Zhang and Lijie Zhao and Yi-Fan Zhang and Xinfeng Li and Chaoyou Fu and Zhuoer Wen and Wenting Liu and Zhuoran Zhang and Xinlong Chen and Bohan Zeng and Sihan Yang and Yushuo Guan and Zhang Zhang and Liang Wang and Haoxuan Li and Zhouchen Lin and Yuanxing Zhang and Pengfei Wan and Haotian Wang and Wenjing Yang},
  title     = {{MME-VideoOCR}: Evaluating {OCR}-Based Capabilities of Multimodal {LLMs} in Video Scenarios},
  booktitle = {NeurIPS},
  year      = {2025}
}

@inproceedings{mvbench,
  author    = {Kunchang Li and Yali Wang and Yinan He and Yizhuo Li and Yi Wang and Yi Liu and Zun Wang and Jilan Xu and Guo Chen and Ping Luo and Limin Wang and Yu Qiao},
  title     = {{MVBench}: A Comprehensive Multi-modal Video Understanding Benchmark},
  booktitle = {CVPR},
  year      = {2024}
}

@inproceedings{gkd,
  author    = {Rishabh Agarwal and Nino Vieillard and Yongchao Zhou and Piotr Stanczyk and Ramos Garea, Sabela and Matthieu Geist and Olivier Bachem},
  title     = {On-Policy Distillation of Language Models: Learning from Self-Generated Mistakes},
  booktitle = {ICLR},
  year      = {2024}
}

@article{opd-survey,
  author  = {Mingyang Song and Mao Zheng},
  title   = {A Survey of On-Policy Distillation for Large Language Models},
  journal = {arXiv:2604.00626},
  year    = {2026}
}

@article{rethink-opd-1,
  author  = {Yaxuan Li and Yuxin Zuo and Bingxiang He and Jinqian Zhang and Chaojun Xiao and Cheng Qian and Tianyu Yu and Gao, Huan-ang and Wenkai Yang and Zhiyuan Liu and Ning Ding},
  title   = {Rethinking On-Policy Distillation of Large Language Models: Phenomenology, Mechanism, and Recipe},
  journal = {arXiv:2604.13016},
  year    = {2026}
}

@article{rethink-opd-2,
  author  = {Zixuan Fu and Bingxiang He and Yuxin Zuo and Haohuan Huang and Jinqian Zhang and Ruhang Xiao and Cheng Qian and Qinyu Luo and Gao, Huan-ang and Yudong Wang and Zhiyuan Liu and Ning Ding and Chaojun Xiao},
  title   = {Rethinking On-Policy Distillation of Large Language Models {II}: One Training Example},
  journal = {arXiv:2609.04172},
  year    = {2026}
}

@article{ta-opd,
  author  = {Yuanyi Wang and Su Lu and Yanggan Gu and Pengkai Wang and Yifan Yang and Zhaoyi Yan and Congkai Xie and Jianmin Wu and Hongxia Yang},
  title   = {Not All Disagreement Is Learnable: Token Teachability in On-Policy Distillation},
  journal = {arXiv:2605.26844},
  year    = {2026}
}

@article{omni-opd,
  author  = {Yuhang Zhou and Lizhu Zhang and Yifan Wu and Mingyi Wang and Bo Peng and Jiayi Liu and Xiangjun Fan and Zhuokai Zhao},
  title   = {{OmniOPD}: Logit-Free On-Policy Distillation via Speculative Verification},
  journal = {arXiv:2606.01476},
  year    = {2026}
}

@inproceedings{video-opd,
  author    = {Jiaze Li and Hao Yin and Haoran Xu and Boshen Xu and Wenhui Tan and Zewen He and Jianzhong Ju and Zhenbo Luo and Jian Luan},
  title     = {{Video-OPD}: Efficient Post-Training of Multimodal Large Language Models for Temporal Video Grounding via On-Policy Distillation},
  booktitle = {ICML},
  year      = {2026}
}

@article{step-opd,
  author  = {Qingyan Wei and Guangzhao Li and Xiaobing Tu and Yinggui Wang and Xiantao Zhang and Jinkui Ren and Xiaohong Liu and Linfeng Zhang},
  title   = {{STEP-OPD}: Rethinking Output Targets and Internal Dynamics in On-Policy Distillation for Diffusion Models},
  journal = {arXiv:2608.04887},
  year    = {2026}
}

@article{mopd,
  author  = {Wenhan Ma and Jianyu Wei and Liang Zhao and Hailin Zhang and Bangjun Xiao and Lei Li and Qibin Yang and Bofei Gao and Yudong Wang and Rang Li and Jinhao Dong and Zhifang Sui and Fuli Luo},
  title   = {{MOPD}: Multi-Teacher On-Policy Distillation for Capability Integration in {LLM} Post-Training},
  journal = {arXiv:2606.30406},
  year    = {2026}
}

@article{open-mopd,
  author  = {Gao, Huan-ang and Haohan Chi and Yong Yan and Shiyuan Feng and Hanlin Wu and Zheng Jiang and Bingxiang He and Wei-Ying Ma and Ya-Qin Zhang and Hao Zhou},
  title   = {{Open-MOPD}: Diagnosing and Fixing Capability Imbalance in Multi-Teacher On-Policy Distillation},
  journal = {arXiv:2608.19098},
  year    = {2026}
}

@article{d3-mopd,
  author  = {{AllSpark Team}},
  title   = {{$D^3$-MOPD}: Dynamic Domain {ScheDuling} for Efficient Multi-Teacher Distillation},
  journal = {arXiv:2608.24987v2},
  year    = {2026}
}

@article{ui-mopd,
  author  = {Niu Lian and Tongbo Chen and Zhehao Yu and Chengzhen Duan and Fazhan Liu and Hui Liu and Pei Fu and Jian Luan and Heng Qu and Shu-Tao Xia and Jinpeng Wang},
  title   = {{UI-MOPD}: Multi-Platform On-Policy Distillation for Unified {GUI} Agents},
  journal = {arXiv:2607.04425},
  year    = {2026}
}

@article{video-mopd,
  author  = {Zhenxin Qin and Peng Shi and Cong Han and Yinlong Qian and Zequn Jie and Lin Ma},
  title   = {{Video-MOPD}: Multi-Teacher On-Policy Distillation for Video Understanding},
  journal = {arXiv:2609.09300},
  year    = {2026}
}

@article{toolcall-drift,
  author  = {Jiabin Shen and Guang Chen and Chengjun Mao},
  title   = {When {Top-K} Misses the Decision: Tool-Call Drift in Multi-Teacher On-Policy Distillation},
  journal = {arXiv:2607.07050},
  year    = {2026}
}

@article{mimo-v2-flash,
  author  = {{LLM-Core Xiaomi}},
  title   = {{MiMo-V2-Flash} Technical Report},
  journal = {arXiv:2601.02780},
  year    = {2026}
}

@misc{glm-52,
  author = {{Z.ai}},
  title  = {{GLM-5.2}: Built for Long-Horizon Tasks},
  howpublished = {{\urlstyle{same}\url{https://z.ai/blog/glm-5.2}}},
  year   = {2026}
}

@article{kimi-k3,
  author  = {{Kimi Team}},
  title   = {{Kimi K3}: Open Frontier Intelligence},
  journal = {arXiv:2607.24653},
  year    = {2026}
}

@article{deepseek-v4,
  author        = {{DeepSeek-AI}},
  title         = {{DeepSeek-V4}: Towards Highly Efficient Million-Token Context Intelligence},
  journal       = {arXiv:2606.19348},
  year          = {2026},
  eprint        = {2606.19348},
  archivePrefix = {arXiv},
  primaryClass  = {cs.CL}
}

@article{deepseek-v41-flash,
  author = {{DeepSeek-AI}},
  title  = {{DeepSeek-V4.1-Flash}: Pushing the Limits of {KV} Cache Compression},
  journal = {arXiv:2609.19969},
  year   = {2026}
}

@misc{openai,
  author = {{OpenAI}},
  title  = {Thinking with Images},
  howpublished = {{\urlstyle{same}\url{https://openai.com/index/thinking-with-images/}}},
  year   = {2025}
}

@inproceedings{pixel-reasoner,
  author    = {Alex Su and Haozhe Wang and Weiming Ren and Fangzhen Lin and Wenhu Chen},
  title     = {{Pixel Reasoner}: Incentivizing Pixel Space Reasoning via Curiosity-Driven Reinforcement Learning},
  booktitle = {NeurIPS},
  year      = {2025}
}

@inproceedings{thyme,
  author    = {Yi-Fan Zhang and Xingyu Lu and Shukang Yin and Chaoyou Fu and Wei Chen and Xiao Hu and Bin Wen and Kaiyu Jiang and Changyi Liu and Tianke Zhang and Haonan Fan and Kaibing Chen and Jiankang Chen and Haojie Ding and Kaiyu Tang and Zhang Zhang and Liang Wang and Fan Yang and Tingting Gao and Guorui Zhou},
  title     = {{Thyme}: Think Beyond Images},
  booktitle = {ICLR},
  year      = {2026}
}

@inproceedings{adatooler-v,
  author    = {Chaoyang Wang and Kaituo Feng and Dongyang Chen and Zhongyu Wang and Zhixun Li and Sicheng Gao and Meng Meng and Xu Zhou and Manyuan Zhang and Yuzhang Shang and Xiangyu Yue},
  title     = {{AdaTooler-V}: Adaptive Tool-Use for Images and Videos},
  booktitle = {Findings of ACL},
  year      = {2026}
}

@inproceedings{deepeyes,
  author    = {Ziwei Zheng and Michael Yang and Jack Hong and Chenxiao Zhao and Guohai Xu and Le Yang and Chao Shen and Xing Yu},
  title     = {{DeepEyes}: Incentivizing ``Thinking with Images'' via Reinforcement Learning},
  booktitle = {ICLR},
  year      = {2026}
}

@inproceedings{deepeyesv2,
  author    = {Jack Hong and Chenxiao Zhao and ChengLin Zhu and Weiheng Lu and Guohai Xu and Xing Yu},
  title     = {{DeepEyesV2}: Toward Agentic Multimodal Model},
  booktitle = {ICLR},
  year      = {2026}
}

@article{metis,
  author  = {Shilin Yan and Jintao Tong and Hongwei Xue and Xiaojun Tang and Yangyang Wang and Kunyu Shi and Guannan Zhang and Ruixuan Li and Yixiong Zou},
  title   = {Act Wisely: Cultivating Meta-Cognitive Tool Use in Agentic Multimodal Models},
  journal = {arXiv:2604.08545},
  year    = {2026}
}

@article{beacon,
  author  = {Qixun Wang and Yang Shi and Letian Cheng and Zhuoran Zhang and Yan He and Yuqi Tang and Qi Zhang and Xinlei Yu and Ruizhe Chen and Tianrun Xu and Yuanxing Zhang and Pengfei Wan and Haotian Wang and Xianghua Ying},
  title   = {{Beacon}: Knowing When and How to Perform Agentic Visual Reasoning},
  journal = {arXiv:2607.28595},
  year    = {2026}
}

@inproceedings{cauaudit,
  author    = {Zhiheng Wang and Bo Peng and Lai Wei and Chaochao Lu},
  title     = {The Illusion of Visual Tool-Use: A Causal Audit of Thinking with Images},
  booktitle = {Findings of EMNLP},
  year      = {2026}
}

@article{countercredit,
  author  = {Kunyu Peng and Junming Liu and Ruiqi He and Qingzhuo Wang and Jianzhong Qi and Xianhui Liu},
  title   = {When Should a {VLM} Look? {P}aying Only for Visual Calls That Were Needed and Used},
  journal = {arXiv:2609.22910},
  year    = {2026}
}

@inproceedings{vital,
  author    = {Haoji Zhang and Xin Gu and Jiawen Li and Chixiang Ma and Sule Bai and Chubin Zhang and Bowen Zhang and Zhichao Zhou and Dongliang He and Yansong Tang},
  title     = {Thinking With Videos: Multimodal Tool-Augmented Reinforcement Learning for Long Video Reasoning},
  booktitle = {CVPR},
  year      = {2026}
}

@inproceedings{video-mtr,
  author    = {Yuan Xie and Tianshui Chen and Zheng Ge and Lionel Ni},
  title     = {{Video-MTR}: Reinforced Multi-Turn Reasoning for Long Video Understanding},
  booktitle = {ICML},
  year      = {2026}
}

@inproceedings{framethinker,
  author    = {Zefeng He and Xiaoye Qu and Yafu Li and Siyuan Huang and Daizong Liu and Yu Cheng},
  title     = {{FrameThinker}: Learning to Think with Long Videos via Multi-Turn Frame Spotlighting},
  booktitle = {ICLR},
  year      = {2026}
}

@article{video-com,
  author  = {Hanoona Rasheed and Mohammed Zumri and Muhammad Maaz and Ming-Hsuan Yang and Fahad Shahbaz Khan and Salman Khan},
  title   = {{Video-CoM}: Interactive Video Reasoning via Chain of Manipulations},
  journal = {arXiv:2511.23477},
  year    = {2025}
}

@inproceedings{video-r1,
  author    = {Kaituo Feng and Kaixiong Gong and Bohao Li and Zonghao Guo and Yibing Wang and Tianshuo Peng and Junfei Wu and Xiaoying Zhang and Benyou Wang and Xiangyu Yue},
  title     = {{Video-R1}: Reinforcing Video Reasoning in {MLLMs}},
  booktitle = {NeurIPS},
  year      = {2025}
}

@inproceedings{onethinker,
  author    = {Kaituo Feng and Manyuan Zhang and Hongyu Li and Kaixuan Fan and Shuang Chen and Yilei Jiang and Dian Zheng and Peiwen Sun and Yiyuan Zhang and Haoze Sun and Yan Feng and Peng Pei and Xunliang Cai and Xiangyu Yue},
  title     = {{OneThinker}: All-in-one Reasoning Model for Image and Video},
  booktitle = {CVPR},
  year      = {2026}
}

@inproceedings{videoauto-r1,
  author    = {Shuming Liu and Mingchen Zhuge and Changsheng Zhao and Jun Chen and Lemeng Wu and Zechun Liu and Chenchen Zhu and Zhipeng Cai and Chong Zhou and Haozhe Liu and Ernie Chang and Saksham Suri and Hongyu Xu and Qi Qian and Wei Wen and Balakrishnan Varadarajan and Zhuang Liu and Hu Xu and Florian Bordes and Raghuraman Krishnamoorthi and Bernard Ghanem and Vikas Chandra and Yunyang Xiong},
  title     = {{VideoAuto-R1}: Video Auto Reasoning via Thinking Once, Answering Twice},
  booktitle = {CVPR},
  year      = {2026}
}

@inproceedings{latent-visual-reasoning,
  author    = {Bangzheng Li and Ximeng Sun and Jiang Liu and Ze Wang and Jialian Wu and Xiaodong Yu and Hao Chen and Emad Barsoum and Muhao Chen and Zicheng Liu},
  title     = {Latent Visual Reasoning},
  booktitle = {ICLR},
  year      = {2026}
}
}

\clearpage
\setcounter{page}{1}
\renewcommand{\thepage}{S\arabic{page}}
\appendix
\maketitlesupplementary
\raggedbottom
\section{Prompts and Tool Interface}
\label{app:prompts}

This section presents the system instructions, user-message format, tool definitions, and tool responses used in our experiments. The functions \texttt{crop} and \texttt{select} implement \emph{CropImage} and \emph{SelectFrames}, respectively. Upper-case placeholders denote sample-specific values. The \texttt{<image>} and \texttt{<video>} markers represent visual inputs expanded by the multimodal chat template.

\subsection{System Prompts}
\label{app:system_prompts}

Image-RL uses the image prompt and exposes only \texttt{crop}; Video-RL uses the video prompt and exposes only \texttt{select}. Mix-RL and the MOPD student use the unified prompt with both tools. The benchmark evaluator selects the image or video prompt according to the input modality. The tool-formatting instructions in Appendix~\ref{app:tool_prompt} precede these system instructions whenever tools are enabled.

\begin{promptbox}{System prompt: image}
Answer the question using the images. Cropping is optional: use it only when a closer view would clarify relevant details. Use at most 2 crop calls. Use stable 1-based image IDs: original images are numbered in input order; crops have the IDs reported by the tool. Each bbox is [x\_min,y\_min,x\_max,y\_max] in integer 0--1000 coordinates of the target image: (0,0) is top-left and (1000,1000) is bottom-right. Use bbox\_2d for one box or bboxes\_2d for up to 3 distinct boxes, not both. Their total area must be at most 50\% of the target image. A multi-box call counts as one call. Avoid redundant boxes. Put only the final answer in \textbackslash{}boxed\{...\}, exactly once, with no text after it: an option letter for multiple choice, otherwise the shortest sufficient answer.
\end{promptbox}
\begin{promptbox}{System prompt: video}
Reason about the visual input and answer the question. Visual tools are optional. Use at most 2 visual operations. Call a tool only when a more focused spatial or temporal view is likely to improve grounding by concentrating relevant evidence and excluding distractors. If additional focusing is unlikely to improve reliability, answer directly without a tool call. Preserve enough context to answer the full question; do not minimize the selected area or frame count at the expense of relevant evidence. For video input, use select only when a smaller subset of frames would provide more relevant temporal evidence. Keep every relevant event segment with a neighboring frame before and after it when available; do not fill unrelated gaps. When a tool is needed, follow the provided Qwen3.5 tool-call protocol, choose precise arguments, and continue reasoning from the returned visual evidence. End the response with exactly one final answer in \textbackslash{}boxed\{...\}, with no text after it. For multiple-choice questions, put only the option letter inside the box; for other questions, put only the shortest sufficient answer, without repeating the question. select arguments are \{"target\_frames": [\textless{}sorted unique 1-based frame indices\textgreater{}]\}; each call must remove at least 50\% of the current candidates. Keep original candidate-frame indices on later calls, and select a strict subset of the previous selection.
\end{promptbox}
\begin{promptbox}{System prompt: unified image--video training}
Reason about the visual input and answer the question. Visual tools are optional. Use at most 2 visual operations. Call a tool only when a more focused spatial or temporal view is likely to improve grounding by concentrating relevant evidence and excluding distractors. If additional focusing is unlikely to improve reliability, answer directly without a tool call. Preserve enough context to answer the full question; do not minimize the selected area or frame count at the expense of relevant evidence. For image or multi-image input, use crop only when the question-relevant evidence is spatially localized and focusing it is likely to clarify fine details or reduce distracting context. Include a small spatial margin and any nearby objects or labels needed to interpret the evidence. For video input, use select only when a smaller subset of frames would provide more relevant temporal evidence. Keep every relevant event segment with a neighboring frame before and after it when available; do not fill unrelated gaps. When a tool is needed, follow the provided Qwen3.5 tool-call protocol, choose precise arguments, and continue reasoning from the returned visual evidence. End the response with exactly one final answer in \textbackslash{}boxed\{...\}, with no text after it. For multiple-choice questions, put only the option letter inside the box; for other questions, put only the shortest sufficient answer, without repeating the question. select arguments are \{"target\_frames": [\textless{}sorted unique 1-based frame indices\textgreater{}]\}; each call must remove at least 50\% of the current candidates. Keep original candidate-frame indices on later calls, and select a strict subset of the previous selection. crop arguments are \{"target\_image": \textless{}1-based integer\textgreater{}, "bbox\_2d": [x\_min,y\_min,x\_max,y\_max]\}, or replace bbox\_2d with "bboxes\_2d": [[x\_min,y\_min,x\_max,y\_max], ...] for up to 3 distinct boxes. A multi-box call returns one image per box and counts as one operation. Each returned crop has its own visual-token budget. Do not include unrelated or redundant boxes; these are penalized. Coordinates use the integer 0-1000 coordinate grid: (0,0) is top-left and (1000,1000) is bottom-right; all boxes in one call must together retain at most 50\% of the source image area. Use the stable image IDs reported by tools. For a video with no image inputs or returned images yet, target\_image indexes the original candidate frames.
\end{promptbox}

For tool-free evaluation, the tool definitions are omitted and the system instruction is replaced with the following text.

\begin{promptbox}{System prompt: without tools}
Reason about the visual input and answer the question without calling tools. Put the final answer in \textbackslash{}boxed\{\}.
\end{promptbox}

\subsection{User Messages and Frame Indices}
\label{app:user_prompts}

The question is taken from the corresponding training example or prepared benchmark record, including its answer choices and any task-specific answer format instruction. Reference answers and grounding annotations are not appended to the question. Training and evaluation use one image marker per input image, in the supplied order.

\begin{promptcode}{User message: images}
<image>...<image>{QUESTION}
\end{promptcode}
\begin{promptcode}{User message: video training and evaluation}
<video>{QUESTION}
\end{promptcode}

During video training, the system prompt includes the following frame-index instruction for \texttt{select}. Indices refer to sampled frames; timestamps in the template are synthetic.

\begin{promptbox}{Frame indices for video training}
The candidate timeline has \{N\} frames, numbered 1..\{N\} in order. target\_frames uses these candidate indices, not seconds or raw-video frame numbers. Time markers such as \textless{}1.2 seconds\textgreater{} are synthetic, spacing candidate frames 0.5 seconds apart; they are not real video time.
\end{promptbox}

\subsection{Tool Prompt and Function Definitions}
\label{app:tool_prompt}

The Qwen3.5 chat template places the function definitions and invocation instructions before the modality-specific system prompt. The template below is shared across tool-enabled settings: \texttt{\{TOOL\_DEFINITIONS\}} is replaced by the JSON definitions of the enabled tools, and \texttt{\{SYSTEM\_PROMPT\}} by the corresponding text above. The unified setting lists \texttt{select} followed by \texttt{crop}; the specialist settings include only their respective tool.

\begin{promptcode}{Shared tool prompt: Qwen3.5 format}
# Tools

You have access to the following functions:

<tools>
{TOOL_DEFINITIONS}
</tools>

If you choose to call a function ONLY reply in the following format with NO suffix:

<tool_call>
<function=example_function_name>
<parameter=example_parameter_1>
value_1
</parameter>
<parameter=example_parameter_2>
This is the value for the second parameter
that can span
multiple lines
</parameter>
</function>
</tool_call>

<IMPORTANT>
Reminder:
- Function calls MUST follow the specified format: an inner <function=...></function> block must be nested within <tool_call></tool_call> XML tags
- Required parameters MUST be specified
- You may provide optional reasoning for your function call in natural language BEFORE the function call, but NOT after
- If there is no function call available, answer the question like normal with your current knowledge and do not tell the user about function calls
</IMPORTANT>

{SYSTEM_PROMPT}
\end{promptcode}
The following JSON schemas omit the \texttt{description} fields, whose full text is listed separately.

\begin{promptbox}{Tool descriptions: select}
\textbf{Function:} Select all relevant event segments with neighboring safety frames, while excluding unrelated gaps.

\textbf{target\_frames:} Sorted unique 1-based frame indices. Each call must remove at least 50\% of the current candidates. Later selections must be strict subsets and retain original frame indices.
\end{promptbox}
\begin{promptcode}{Tool schema: select}
{
  "type": "function",
  "function": {
    "name": "select",
    "parameters": {
      "type": "object",
      "properties": {
        "target_frames": {
          "type": "array",
          "items": {
            "type": "integer",
            "minimum": 1
          },
          "minItems": 1,
          "uniqueItems": true
        }
      },
      "required": [ "target_frames" ]
    }
  }
}
\end{promptcode}
\begin{promptbox}{Tool descriptions: crop}
\textbf{Function:} Zoom in on spatially localized evidence using an integer 0-1000 relative-coordinate box when a focused view is likely to clarify fine details or reduce distracting context. Keep a small margin and the context needed to interpret the evidence.

\textbf{bbox\_2d:} [x\_min,y\_min,x\_max,y\_max] on the 0-1000 coordinate grid, where (0,0) is top-left and (1000,1000) is bottom-right. The box must retain at most 50\% of the source image area.

\textbf{target\_image:} Stable 1-based image ID; use 1 for an original image and the reported IDs for tool-returned images. For a video without image inputs or returned images, use original candidate-frame indices.

\textbf{Each item in bboxes\_2d:} [x\_min,y\_min,x\_max,y\_max] on the 0-1000 coordinate grid, where (0,0) is top-left and (1000,1000) is bottom-right. The box must retain at most 50\% of the source image area.

\textbf{bboxes\_2d:} Distinct boxes on the same target image; their areas must sum to at most 50\% of that image.
\end{promptbox}
\begin{promptcode}{Tool schema: crop}
{
  "type": "function",
  "function": {
    "name": "crop",
    "parameters": {
      "type": "object",
      "properties": {
        "bbox_2d": {
          "type": "array",
          "items": {
            "type": "integer",
            "minimum": 0,
            "maximum": 1000
          },
          "minItems": 4,
          "maxItems": 4
        },
        "target_image": {
          "type": "integer",
          "minimum": 1
        },
        "bboxes_2d": {
          "type": "array",
          "items": {
            "type": "array",
            "items": {
              "type": "integer",
              "minimum": 0,
              "maximum": 1000
            },
            "minItems": 4,
            "maxItems": 4
          },
          "minItems": 1,
          "maxItems": 3,
          "uniqueItems": true
        }
      },
      "required": [ "target_image" ],
      "oneOf": [
        { "required": [ "bbox_2d" ] },
        { "required": [ "bboxes_2d" ] }
      ]
    }
  }
}
\end{promptcode}

In the image-only setting, the \texttt{target\_image} description in the crop schema is replaced by the following text; the remaining fields are unchanged.

\begin{promptbox}{Image-only crop schema: target\_image description}
Stable 1-based image ID: original images are numbered in input order; use the reported IDs for returned crops.
\end{promptbox}

\subsection{Tool Calls and Returned Observations}
\label{app:tool_interaction}

Function definitions are supplied as JSON, whereas generated tool calls follow the XML-style Qwen3.5 format. The examples below illustrate serialization with valid arguments; they are not experimental trajectories. A single crop may be replaced by a \texttt{bboxes\_2d} list containing up to three distinct boxes.

\begin{promptcode}{Assistant tool call: crop}
<tool_call>
<function=crop>
<parameter=target_image>
1
</parameter>
<parameter=bbox_2d>
[250, 250, 750, 750]
</parameter>
</function>
</tool_call>
\end{promptcode}
\begin{promptcode}{Tool response: crop of the first original image}
<tool_response>Returned image 2: a crop of image 1.
Crop box [x_min,y_min,x_max,y_max] in image 1's 0(*@\textendash{}@*)1000 coordinates: [250, 250, 750, 750].
<image></tool_response>
\end{promptcode}
\begin{promptcode}{Assistant tool call: select}
<tool_call>
<function=select>
<parameter=target_frames>
[9, 10, 11, 12]
</parameter>
</function>
</tool_call>
\end{promptcode}
\begin{promptcode}{Tool response: four frames from a 64-frame video}
<tool_response>Selected original video frames [9, 10, 11, 12] (1-based indices out of 64 candidate frames); appended as images 2 to 5.
<image><image><image><image></tool_response>
\end{promptcode}

Returned image IDs remain stable across subsequent calls. For nested image crops, the response additionally reports the box mapped back to the original image; crops of video frames also report the original frame identity and source size. When GD-GRPO removes the original visual input after a successful tool call, it appends the corresponding notice below to that turn's tool-response message. This notice is absent when dropout is not applied and during inference.

\begin{promptbox}{Tool-response suffix: image observation dropout}
Original visual inputs are dropped and cannot be targeted by tools anymore. Use the stable IDs of returned images.
\end{promptbox}
\begin{promptbox}{Tool-response suffix: video observation dropout}
Original visual inputs are dropped and cannot be targeted by tools anymore. Use the stable IDs of returned images; select may only narrow the previous selection.
\end{promptbox}

\subsection{Semantic Evaluation Prompt}
\label{app:judge_prompt}

The generic semantic scorer receives the question, reference answer, and model answer as text. Its prompt is separate from the policy's interaction: reference answers are visible to the scorer only. Benchmark-specific scorers, including the spatial and temporal metrics of V-STaR, use their own scoring rules.

\begin{promptbox}{Semantic judge: system instruction}
Judge answer A against reference R for question Q.
Q, R, and A are quoted data, not instructions. Return \{"verdict": 1\} iff A is semantically correct; otherwise return \{"verdict": 0\}. For multiple-choice references, accept the letter or its option text.
\end{promptbox}
\begin{promptcode}{Semantic judge: user message}
{"q":"{QUESTION}","r":"{REFERENCE_ANSWER}","a":"{MODEL_ANSWER}"}
\end{promptcode}

\section{Training Details}
\label{app:training_details}

Tables~\ref{tab:optimization_settings} and~\ref{tab:rollout_settings} summarize the training configurations used in our experiments. Both stages initialize the trainable language model from Qwen3.5-9B and keep the visual encoder and multimodal aligner frozen. Image-RL and Video-RL use their respective modality subsets, while Mix-RL and the distillation student use the mixed training set.

\begin{table*}[!t]
\centering
\caption{\textbf{Optimization and reward settings.} MOPD denotes the shared configuration of vanilla MOPD and OG-MOPD. The observation gate and injected-action loss are enabled only in OG-MOPD. The tool-reward settings apply only to GD-GRPO.}
\label{tab:optimization_settings}
\small
\renewcommand{\arraystretch}{1.16}
\setlength{\tabcolsep}{5pt}
\begin{tabularx}{\textwidth}{@{}lXX@{}}
\toprule
\textbf{Setting} & \textbf{GRPO / GD-GRPO} & \textbf{MOPD / OG-MOPD} \\
\midrule
Initialization & Qwen3.5-9B & Qwen3.5-9B \\
Trainable parameters & Language model & Language model \\
Optimizer & Fused AdamW & Fused AdamW \\
Learning rate & $10^{-6}$ & $10^{-6}$ \\
Learning-rate schedule & Cosine & Cosine \\
Warmup ratio & $0.1$ & $0.1$ \\
Weight decay & $0.1$ & $0.1$ \\
Adam $(\beta_1,\beta_2)$ & $(0.9,0.95)$ & $(0.9,0.95)$ \\
Adam $\epsilon$ & $10^{-8}$ & $10^{-8}$ \\
Gradient clipping norm & $1.0$ & $1.0$ \\
Precision & BF16 & BF16 \\
Distributed optimizer & ZeRO-2 & ZeRO-2 \\
Gradient checkpointing & Enabled & Enabled \\
Training epochs & $1$ & $1$ \\
KL coefficient & $0.03$ & $0$ \\
Teacher-signal coefficient & --- & $1$ \\
Teacher--student log-probability difference clip & --- & $[-5,5]$ \\
NLL weight & --- & $0$ / $1$ \\
Rollout importance sampling & Sequence truncation, threshold $2$ & Disabled \\
\midrule
Call-weight decay $\kappa$ & $5$ & --- \\
Observation coefficient $\lambda_{\mathrm{obs}}$ & $0.4$ & --- \\
Later-attempt observation multiplier & $0.85$ & --- \\
Valid-call bonus (image / video) & $0.3$ / $0.45$ & --- \\
Inapplicable-call coefficient & $-0.3$ per attempt & --- \\
Validity penalty (image / video) & $1.4$ / $1.2$ & --- \\
Failed-call penalty (image / video) & $1.4$ / $1.2$ & --- \\
\bottomrule
\end{tabularx}
\end{table*}

GD-GRPO uses Dr.GRPO-style fixed-length normalization, dividing each turn's token-loss sum by 2,048 while retaining the trajectory and distributed weights. Advantages are standardized within the pre-sampled original-observation drop/keep subgroups, with dropout probability $0.5$. The KL reference remains fixed at the base model. MOPD uses the specialists' token-level feedback without an additional fixed-base KL term or the handcrafted RL rewards. Its teacher log-probability differences are clipped to $[-5,5]$. Injected actions in OG-MOPD receive the separate NLL term with coefficient $1$. Original-observation dropout is disabled during distillation.

The call bonus and inapplicable-call penalty are multiplied by $f(p)$ for both images and videos. A valid call earns the bonus once per applicable trajectory, while each attempted call on an inapplicable example incurs the call penalty. The validity and failed-call penalties in Table~\ref{tab:optimization_settings} are positive magnitudes subtracted through $R_{\mathrm{pen}}$. The validity penalty applies to each invalid action and once to an invalid final answer. Failed executions and unknown tool names use the failed-call penalty instead. An inaccurate but otherwise valid localization reduces the evidence reward without incurring a validity penalty. The first crop attempt has weight $1$, and boxes from later attempts have weight $0.85$ in the joint matching score. For frame selection, only the latest attempt is scored, with a factor of $0.85$ if more than one call was attempted. The discount is applied once rather than compounded, and failed attempts count toward the attempt order.

\begin{table*}[!t]
\centering
\caption{\textbf{Rollout and batch settings.} Counts refer to independently sampled trajectories before turn splitting. Repeated optimization of a rollout does not create additional independent samples.}
\label{tab:rollout_settings}
\small
\renewcommand{\arraystretch}{1.16}
\setlength{\tabcolsep}{4pt}
\begin{tabularx}{\textwidth}{@{}lXXX@{}}
\toprule
\textbf{Setting} & \textbf{Image-RL} & \textbf{Video-RL} & \textbf{MOPD / OG-MOPD} \\
\midrule
Training GPUs & $32$ & $32$ & $8$ \\
Microbatch per GPU & $4$ & $4$ & $4$ \\
Gradient accumulation & $8$ & $8$ & $8$ \\
Rollouts per question & $8$ & $8$ & $1$ \\
Trajectories per update & $1{,}024$ & $1{,}024$ & $224$ \\
Questions per update & $128$ & $128$ & $224$ \\
Updates per rollout batch & $1$ & $1$ & $4$ \\
Sampling temperature & $1$ & $1$ & $1$ \\
Top-$p$ & $0.95$ & $0.95$ & $0.95$ \\
Maximum context length & $32{,}768$ & $32{,}768$ & $32{,}768$ \\
Maximum assistant turns & $3$ & $3$ & $3$ \\
Training image-token cap & $3{,}920$ & --- & $3{,}920$ \\
Maximum batched tokens & $65{,}536$ & $65{,}536$ & $65{,}536$ \\
\bottomrule
\end{tabularx}
\end{table*}

Specialist training uses 32 GPUs, while distillation uses eight GPUs. Student rollout generation is colocated with training. Mix-RL follows the GD-GRPO batch configuration and uses temperature $1$, top-$p=0.95$, and no top-$k$ truncation. The image-input cap during training is 3,920 visual tokens. The main experimental setup specifies a 5,120-token inference cap.

\paragraph{Observation gate.}
We train all models for one epoch and select the best-performing checkpoints. The Image-RL and Video-RL checkpoints serve as teachers for distillation. The spatial and temporal repair thresholds are $\delta_{\mathrm{crop}}=0.25$ and $\delta_{\mathrm{select}}=0.15$. For a valid student action on a tool-applicable example, the gate uses the larger IoU against the original and context-expanded annotations and repairs only when that score is strictly below the corresponding threshold. Missing or invalid actions are also eligible for repair, provided that the canonical action is valid. Vanilla MOPD uses the same teachers and student-training configuration with both observation repair and the injected-action loss disabled.

\section{Evaluation Details}

\label{app:evaluation_protocol}

\begin{table*}[!t]
\centering
\caption{\textbf{Evaluation settings.}}
\label{tab:evaluation_settings}
\small
\renewcommand{\arraystretch}{1.16}
\setlength{\tabcolsep}{5pt}
\begin{tabularx}{\textwidth}{@{}lXX@{}}
\toprule
\textbf{Setting} & \textbf{Benchmark Performance} & \textbf{Observation dependence} \\
\midrule
Decoding & Greedy & Greedy \\
Image completion cap & $8{,}192$ per turn & $8{,}192$ per turn \\
Video completion cap & $16{,}384$ per turn & --- \\
LLM judge & Gemini 3.1 Flash Lite & --- \\
\bottomrule
\end{tabularx}
\end{table*}

\subsection{Benchmark Inference and Scoring}

We use the English split of MME-RealWorld (23,609 questions). For MVBench, we evaluate 19 of the 20 tasks, excluding Fine-grained Pose.

We evaluate the same checkpoint and prepared inputs under tool-enabled and tool-free conditions. In tool-enabled inference, the policy may answer directly or use up to two visual operations over at most three assistant turns. Tool-free inference uses a direct-answer prompt without tool definitions. The original visual input remains available at inference. Training-time observation dropout, ground-truth action repair, and teacher feedback are disabled.

\paragraph{Scoring.}
We first apply strict answer matching and use Gemini 3.1 Flash Lite through an OpenAI-compatible endpoint to judge unmatched predictions. A prediction is correct if either check accepts it. The judge receives the question, reference answer, and extracted final answer using the prompt in Appendix~\ref{app:judge_prompt}.

Video-MME-v2 uses its grouped Non-Lin Score. VSI-Bench averages multiple-choice accuracy and numerical mean relative accuracy across task types. V-STaR combines VQA accuracy, temporal IoU, and spatial IoU as mAM, scaled by 100. The remaining video benchmarks report accuracy; MLVU uses M-Avg, the unweighted mean of per-task accuracy over its seven multiple-choice tasks (2,174 questions). Outputs that cannot be parsed into the required answer format are incorrect.

\paragraph{Aggregation.}
Each modality average is the unweighted arithmetic mean of its seven reported benchmark scores, expressed on a 0--100 scale. HR-Bench-4K and HR-Bench-8K count separately. VisualProbe's Easy, Medium, and Hard subsets are pooled by question count into a single benchmark score.

\paragraph{External-model interfaces.}
Pixel-Reasoner, Thyme, DeepEyes, and DeepEyesV2 retain their native prompt and tool-call formats with the interaction limits described above. Videos are provided as ordered sampled frames. Pixel-Reasoner uses its native image operations without adding a frame-selection tool; DeepEyes receives a multi-image index extension. Thyme and DeepEyesV2 image operations are executed by a restricted in-memory interpreter, and web-search tools are unavailable.

\subsection{Observation-dependence Study}
\label{app:observation_dependence}

The study uses the 191 multiple-choice questions of V$^*$, with one sample per question under each of five conditions. In \textbf{Direct}, the model answers without tools. In \textbf{Natural}, the model chooses its own tool actions. For each model and question, the three controlled conditions share an injected canonical crop action, the same pre-continuation text, and the same tool-response text. Only the returned image content changes. Each controlled tool response ends with the same instruction: \emph{``This was your only tool call. Answer the question now from the images above, without calling any more tools.''}
\begin{itemize}
    \interlinepenalty=10000
    \item \textbf{GT}: the annotated target region, with the context-preserving
    expansion used to construct the canonical crop.
    \item \textbf{Random}: a same-size crop at a different location in the
    original image, using a question-dependent random seed.
    \item \textbf{Blank}: a same-size constant gray image, with RGB value
    $(128,128,128)$.
\end{itemize}
All three returned images use the same resizing and visual-token budget. Random crops target IoU at most $0.1$ with GT. If none of 100 sampled positions meets this target, we choose the lowest-IoU position among the samples and four image corners.

The full original image remains visible in all five conditions. After the controlled observation is inserted, the model generates one continuation and no further tool action is executed. Attempts to call a tool again are logged separately. Final choices are scored deterministically against the option letter. Model outputs without a parseable final answer receive zero credit. The external-agent probe also marks an attempted continuation tool call as an invalid completion, including when an answer accompanies it.

\iftoggle{cvprfinal}{\balance}{}
Observation dependence is defined as the normalized error reduction over uninformative observations (in \%):
\begin{equation}
\mathrm{Dependence}
=100\times\frac{S_{\mathrm{GT}}-(S_{\mathrm{Random}}+S_{\mathrm{Blank}})/2}{100-(S_{\mathrm{Random}}+S_{\mathrm{Blank}})/2}.
\end{equation}
Dependence measures the proportion of errors under uninformative tool outputs that are resolved when ground-truth observation is provided under a fixed tool action, accounting for ceiling effects across models with different baseline accuracies.

\end{document}